\pdfoutput=1
\PassOptionsToPackage{usenames,dvipsnames}{xcolor}
\documentclass[11pt]{article}

\usepackage[]{ACL2023}

\usepackage{times}
\usepackage{latexsym}

\usepackage[T1]{fontenc}

\usepackage[utf8]{inputenc}

\usepackage{microtype}

\usepackage{inconsolata}

\usepackage{amsmath,amssymb,amsfonts}
\usepackage{enumerate}
\usepackage{graphicx}
\usepackage{hyperref}
\usepackage[dvipsnames]{xcolor}

\usepackage{booktabs}
\usepackage{multirow,bigstrut}
\usepackage{tabu}
\usepackage{multirow, multicol}
\usepackage{adjustbox}
\usepackage{float}
\usepackage{placeins}
\usepackage{caption}
\usepackage{verbatim}
\usepackage{multirow, makecell}
\usepackage{tabularx}
    \newcolumntype{L}{>{\raggedright\arraybackslash}X}

\title{cTBLS: Augmenting Large Language Models with Conversational Tables}

\author{Anirudh S Sundar, Larry Heck \\
  AI Virtual Assistant (AVA) Lab \\ The Georgia Institute of Technology\\
  \texttt{ \{asundar34,larryheck\}@gatech.edu} \\}

\begin{document}
\maketitle
\begin{abstract}

Optimizing accuracy and performance while eliminating hallucinations of open-domain conversational large language models (LLMs) is an open research challenge. A particularly promising direction is to augment and ground LLMs with information from structured sources. This paper introduces Conversational 
Tables (\mbox{cTBLS}), a three-step architecture to retrieve and generate dialogue responses grounded on retrieved tabular information. \mbox{cTBLS} uses Transformer encoder embeddings for Dense Table Retrieval and obtains up to 125\% relative improvement over the retriever in the previous state-of-the-art system on the \textsc{HyrbiDialogue} dataset. \mbox{cTBLS} then uses a shared process between encoder and decoder models to perform a coarse+fine tabular knowledge (e.g., cell) ranking combined with a GPT-3.5 LLM response generator to yield a 2x relative improvement in ROUGE scores. Finally, human evaluators prefer cTBLs +80\% of the time (coherency, fluency) and judge informativeness to be 4x better than the previous state-of-the-art. 

\end{abstract}
\section{Introduction}
Equipping conversational AI with multimodal capabilities broadens the range of dialogues that humans have with such systems. A persisting challenge in multimodal conversational AI is the development of systems that produce conversationally coherent responses grounded in textual and non-textual modalities \cite{sundar-heck-2022-multimodal}.

It is well-established that large language models (LLMs) possess real-world knowledge stored within their parameters, as demonstrated by recent research \cite{roberts-etal-2020-much, heinzerling-inui-2021-language}. Nevertheless, the incorporation of conversation-specific extrinsic knowledge into these models to yield precise responses remains an active area of investigation.  While humans can easily retrieve contextual information from tables by examining rows and columns, LLMs often struggle to identify relevant information amidst conversational distractions. 

\textsc{HybriDialogue} \cite{nakamura-etal-2022-hybridialogue}, a dataset of conversations grounded on structured and unstructured knowledge from tables and text, introduces the task of responding to messages by utilizing information from external knowledge and prior dialogue turns. The authors also  present an approach and experimental results on \textsc{HybriDialogue} that represents the current state-of-the-art (SoTA).

This paper proposes an extension to the SoTA approach of \textsc{HybriDialogue} in the form of Conversational Tables (\mbox{cTBLS}) \footnote{Our code will be available at \url{https://github.com/avalab-gt/cTBLS}}, a novel three-step encoder-decoder architecture designed to augment LLMs with tabular data in conversational settings.
In the first step, \mbox{cTBLS} uses a dual-encoder Transformer-based  \cite{vaswani2017attention} Dense Table Retriever (DTR) to retrieve the correct table from the entire corpus based on the user's query. The second step employs a fine-tuned dual-encoder Transformer to track system state and rank cells in the retrieved table according to their relevance to the conversation. Finally, \mbox{cTBLS} utilizes GPT-3.5 to generate a natural language response by prompting it with the ranked cells. 

While previous research separated knowledge retrieval and response generation between encoder and decoder models, this paper demonstrates that LLM decoders can perform these tasks jointly when prompted with knowledge sources ranked by language model encoders. Furthermore, by pre-training the Dense Table Retriever to perform retrieval over a corpus of tables, \mbox{cTBLS} can be extended to new knowledge sources without re-training, by appending additional knowledge to the corpus.

\begin{figure*}[t]
    \centering
    \includegraphics[width=\textwidth]{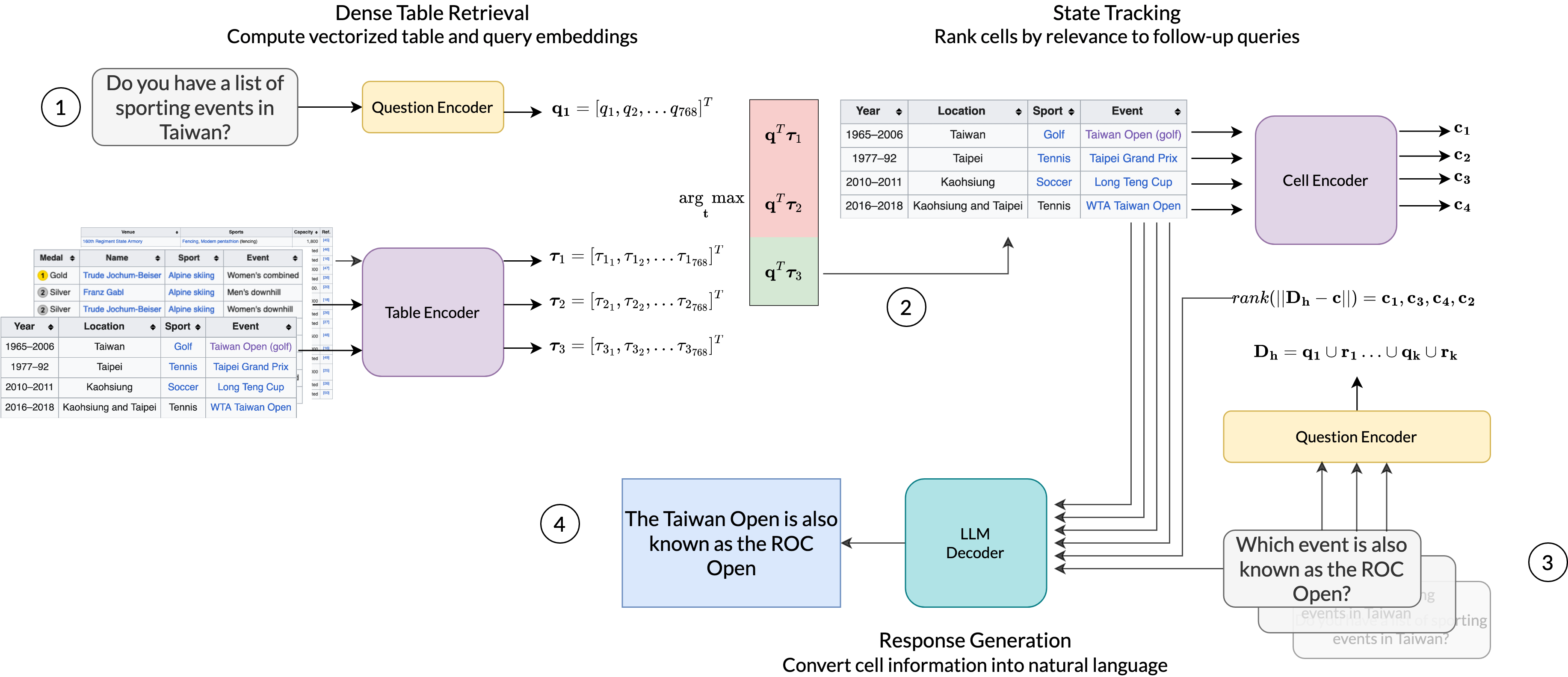}
    \caption{\mbox{cTBLS} for conversations on \textsc{HybriDialogue}. Dense Table Retrieval identifies the table most relevant to the initial query. The retrieved table is provided to the state tracker for follow-up queries. State Tracking ranks cells in the table based on their ability to answer a follow-up query. Response Generation utilizes a LLM Decoder provided with the ranked cell information and the follow-up query to convert tabular data into a natural language response and continue the conversation. Details on individual components are provided in Section \ref{sec:Method}.}
    \label{fig:system_overview}
\end{figure*}

Compared to the previous SoTA, experiments on \mbox{cTBLS} show up to 125\% relative improvement in table retrieval and a 2x relative improvement in ROUGE scores. In addition, human evaluators prefer cTBLs +80\% of the time (coherency, fluency) and judge informativeness to be 4x better than the previous SoTA.


Our contributions are as follows:

 \begin{enumerate}
    \item The introduction of Conversational Tables (\mbox{cTBLS}), a  novel three-step encoder-decoder architecture designed to augment LLMs with tabular data in conversational settings.
     \item Experimental results demonstrating that Dense Table Retrieval, which utilizes neural models fine-tuned with a summary of tabular information, outperforms sparse techniques based on keyword matching for table retrieval.
     \item The presentation of evidence that augmenting state-of-the-art LLM decoders using knowledge sources ranked by encoder language models leads to better results on automatic (ROUGE-Precision) and human (Coherence, Fluency, and Informativeness) evaluation for knowledge-grounded response generation while limiting the number of API calls to these models. 
 \end{enumerate}

This paper presents the \mbox{cTBLS} system and demonstrates its application to the \textsc{HybriDialogue} dataset. In Section \ref{sec:related_work}, we review the existing literature in the fields of Table Question Answering and Knowledge Grounded Response Generation. Section \ref{sec:Method} describes the various components of \mbox{cTBLS} as presented in Figure \ref{fig:system_overview}. In Section~\ref{sec:Experiments}, we evaluate the performance of \mbox{cTBLS} against previous methods for conversations over tables and report experimental results from automatic and human evaluations. Finally, Section \ref{sec:Conclusion} concludes the paper and outlines potential directions for future research.

\section{Related Work}
\label{sec:related_work}
\subsection{Table Question Answering}
Table Question Answering is a well-researched precursor to conversations over tables.  In \textsc{WikiTableQuestions}, \citet{pasupat-liang-2015-compositional} transform HTML tables into a knowledge graph and retrieve the correct answer by converting natural language questions into graph queries. FRETS \cite{jauhar-etal-2016-tables} uses a log-linear model conditioned on alignment scores between cells in tables and individual QA pairs in the training set. \citet{cho2018adversarial} introduce \textsc{NeOp}, a multi-layer sequential network with attention supervision to answer queries conditioned on tables. \citet{hannan2020manymodalqa} propose \textsc{ManyModalQA}, which uses a modality selection network and pre-trained text-based QA, Table-based QA, and Image-based QA models to jointly answer questions over text, tables, and images. \citet{chen-etal-2020-hybridqa} present \textsc{Hybrider}, which performs multi-hop QA over tables using keyword-matching for cell linking followed by BERT \cite{devlin-etal-2019-bert} for reasoning. \citet{chen2020open} propose OTT-QA, which uses a fusion retriever to identify relevant tables and text and a cross-block reader based on a long-range Sparse Attention Transformer \cite{ainslie-etal-2020-etc} to choose the correct answer. \citet{heck2020zero} perform multi-task fine-tuning of Transformer encoders by modeling slot filling as question answering over tabular and visual information in Visual Slot. \citet{herzig-etal-2020-tapas} and \citet{yin-etal-2020-tabert} extend BERT for Table Question Answering by pre-training a masked language model over text-table pairs in \textsc{TaPaS} and TaBERT, respectively. Recent work building off the Transformer architecture for Table Question Answering includes \cite{eisenschlos2021mate, li-etal-2021-dual, herzig-etal-2021-open, zayats-etal-2021-representations, zhao-etal-2022-multihiertt, huang2022mixed, yang2022tableformer,chen2022large}. \citet{jin2022surveytable} provide a comprehensive survey of advancements in Table Question Answering.

\subsection{Knowledge Grounded Response Generation}
Early work related to grounding responses generated by language models in real-world knowledge was motivated by the need to improve prior information for open-domain dialogue \cite{heck2013multimodal, Tur2014eyegaze, hakkani-tr2014probabilistic, huang2015leveraging, jia2017learning}.  More recently,  knowledge grounded response generation has been applied to mitigate the hallucination problem \cite{maynez-etal-2020-faithfulness, shuster-etal-2021-retrieval-augmentation} in LLMs. RAG \cite{lewis2020retrieval} fine-tunes LLMs using Dense Passage Retrieval \cite{karpukhin-etal-2020-dense} over a Wikipedia dump to ground responses for Open Domain Question Answering. KGPT \cite{chen2020kgpt} and SKILL \cite{moiseev-etal-2022-skill} pre-train a Transformer encoder \cite{vaswani2017attention} with English Wikidump for Natural Language Generation. Fusion-in-Decoder \cite{izacard-grave-2021-leveraging}  fine-tunes decoder models using evidence acquired through Dense Passage Retrieval.   

Recent research also includes a dual-stage approach where LLMs generate knowledge sources based on prompts \cite{yu2022generate, bonifacio2022inpars, jeronymo2023inpars}. Closest to our work, Wizard of Wikipedia \cite{dinan2018wizard} jointly optimizes an encoder-decoder Transformer to produce dialogue responses conditioned on retrieved knowledge and dialogue context but does not extend their approach to the multiple modalities. \textsc{RePlug} \cite{shi2023replug} ensembles output responses generated by prompting large language models with inputs from a dense retriever in a zero-shot setting. However, this requires multiple API calls to state-of-the-art LLMs. \textsc{LLM-Augmenter} \cite{peng2023check} incorporates external knowledge in LLM responses by matching keywords in dialogue state to candidate knowledge sources obtained through web-search.  A survey of knowledge fusion in LLMs is available in \citet{colon2021combining} and \citet{richardson2023commonsense}.

In contrast to prior research that focuses on either Table Question Answering or Knowledge Grounded Response Generation, our work, \mbox{cTBLS}, addresses the challenge of generating responses grounded on tabular knowledge. Moreover, while \mbox{cTBLS} is fine-tuned to retrieve tables and filter out incorrect references, it leverages the power of SoTA pre-trained LLMs for response generation. Furthermore, by fine-tuning open-source table and knowledge retrievers to remove inaccurate references, \mbox{cTBLS} reduces the number of API calls to the SoTA LLMs.  



\section{Method}
\label{sec:Method}
The challenge of developing conversational systems grounded in tabular information consists of three tasks, namely table retrieval, system state tracking, and response generation. Table retrieval requires identifying the most relevant table in the dataset based on a given natural language query. System state tracking is responsible for ranking the cells in the table, enabling the system to provide responses to follow-up queries about the table. Finally, response generation involves converting the ranked cells into a natural language response.

\subsection{Table Retrieval}
\label{subsec:tab_ret}
Table retrieval is a prerequisite to answering queries when the exact table to converse over is unspecified. The objective is to identify the correct table from a vast corpus. 
\mbox{cTBLS} proposes formulating table retrieval as document retrieval by assigning a relevance score to each table based on its relevance to the natural language query.  Inspired by \citet{karpukhin-etal-2020-dense} and \citet{huang2013learning}, \mbox{cTBLS} uses a dual-encoder-based Dense Table Retrieval (DTR) model. The DTR model pre-computes a vectorized embedding of all tables in the corpus. Given a query at inference, the retrieved table is closest to the query in the embedded space, indicated by the upper-left portion of Figure \ref{fig:system_overview}. 

\begin{figure}[t]
    \centering
    \includegraphics[width=0.4\textwidth]{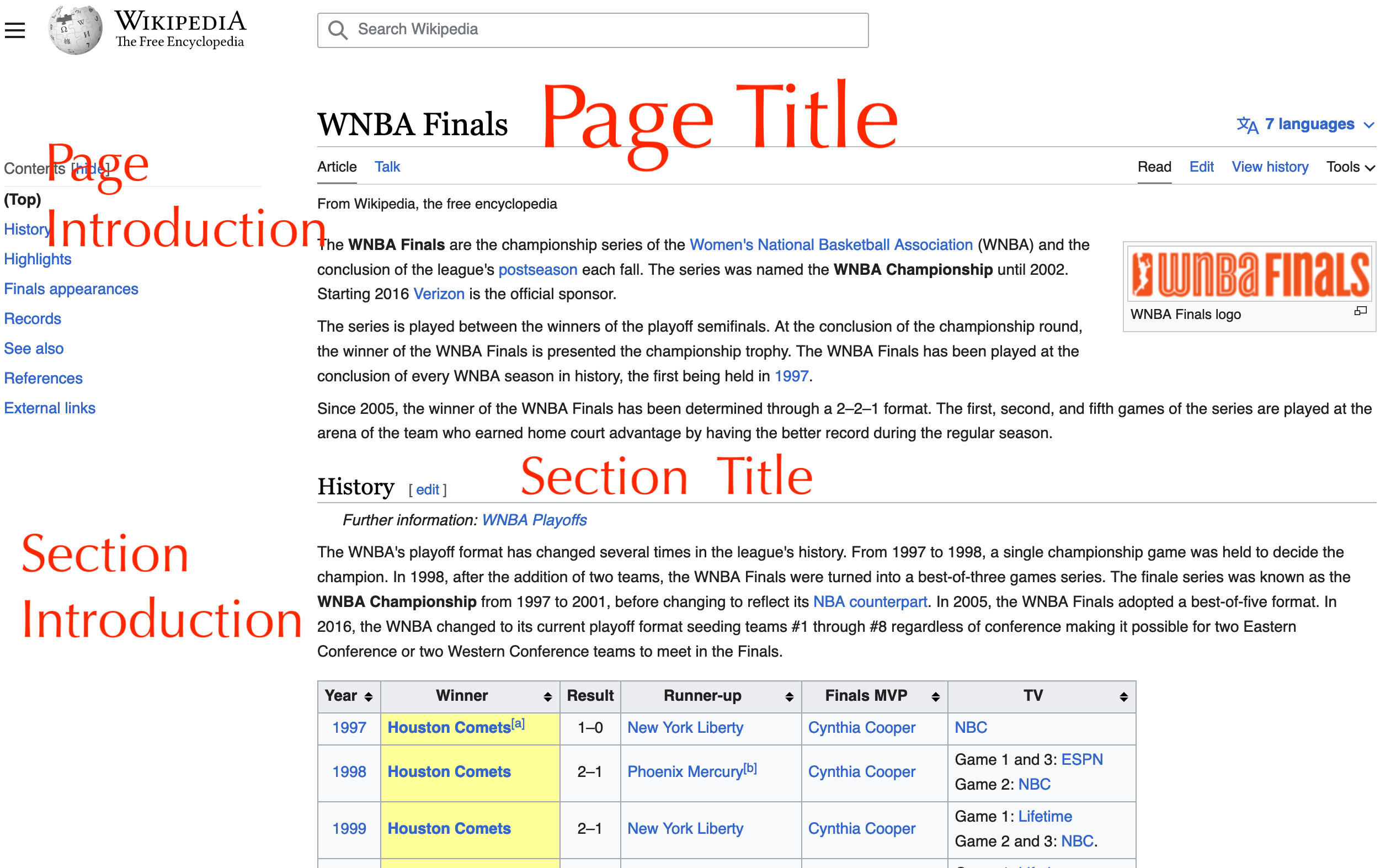}
    \caption{An example of table-associated text in the context of Wikipedia, where the input to the DTR text-encoder includes the page title, the introduction to the article, the section title, and the introduction paragraph.}
    \label{fig:WNBA}
\end{figure}

The DTR model consists of a table encoder and a question encoder, initialized from RoBERTa-base \cite{liu2019roberta}. The input to the table encoder comprises the table's title and, if available, textual information associated with the table. 
Figure \ref{fig:WNBA} presents an example of table-associated text in the context of Wikipedia, where introductions from the page and section provide additional grounding. The input to the question encoder is the current query to be answered. 
Taking the average over the sequence of the last hidden state at the table and question encoder results in 768-dimensional embeddings of the table information and the query. 

The DTR model is optimized through a contrastive prediction task, which aims to maximize the similarity between embeddings of a given query~$q$ and the table to be retrieved $\tau$ while minimizing the similarity to other incorrect tables $\tau_{{n_i}}$ for $i=1,\ldots,N$. As per \cite{karpukhin-etal-2020-dense}, normalized embedding vectors are utilized to optimize the objective in Equation \ref{eq:table_ret}:
\begin{equation}
\small{
\arg\min\limits_
{
\tau\;\;\;\;\;\;\;\;
} 
\left(
-\log 
\frac{
e^{{q}\cdot \tau}
}
{
e^{{q}\cdot \tau}
\;+\;
\sum_{i=1}^N e^{{q}\cdot \tau_{n_i}}
}
\right)
 \label{eq:table_ret}
 }
\end{equation}

Given a batch $B$ of $d$-dimensional query embeddings $\mathbf{Q}$ and table embeddings $\mathbf{T}$, the DTR model computes the similarity $\mathbf{Q}\mathbf{T}^T ( \in \mathbb{R}^{B \times B} )$ between every query and table in the batch. This similarity computation enables the sampling of negatives from other query-table pairs, resulting in $B^2$ training samples in each batch, consisting of $B$ positive pairs along the diagonal and $B^2-B$ negatives.

\subsection{Coarse System State Tracking}
Given a table, system state tracking involves ranking cells in the table by their relevance to conversational queries. In contrast to quesiton-answering, conversational queries require leveraging information from external modalities in conjunction with prior dialogue turns to generate coherent responses \cite{sundar-heck-2022-multimodal}. \mbox{cTBLS} addresses system state tracking through two sub-tasks - coarse and fine system state tracking. Coarse system state tracking ranks cells in the table, while fine system state tracking identifies fine-grained information in the most relevant cell to answer the query. 

\mbox{cTBLS} uses a RoBERTa-base dual-encoder architecture for coarse system state tracking. The cell encoder embeds all cells and associated hyperlinked information, and the question encoder generates embeddings for the dialogue history ($\mathbf{D}_\mathbf{h}$) that includes the current turn's query as well as previous queries and responses.

To rank cells based on their relevance to the follow-up query, as illustrated in the upper-right section of Figure \ref{fig:system_overview}, the question and cell encoders are optimized using a triplet loss configuration. This optimization aims to minimize the distance between the anchor $\mathbf{D}_\mathbf{h}$ and the positive cell $c$, while pushing the negative cell $\overline{c}$ further away from $\mathbf{D}_\mathbf{h}$ by a margin $m$ (Equation~\ref{eq:triplet_loss}).
\begin{equation}
\small{
\arg\min\limits_
{
c_i\;\;\;\;\;\;\;\;\;
} 
(\max
\{
d({\bf D_h},c) - d({\bf D_h},\overline{c}) + m,0
\})
\label{eq:triplet_loss}
}
\end{equation}
\begin{equation}
\small{
    d(x,y) = ||x - y||_2
    }
\end{equation}

For our approach, we utilize an anchor-positive-negative triplet consisting of the complete dialogue history (including queries and responses from previous turns) concatenated with the current query as the anchor, the correct cell as the positive, and other cells from the same table that are not relevant to the query as negatives. We measure the distance between the anchor and the positive and between the anchor and the negatives using the 2-norm distance function $d(\cdot)$.

\subsection{Fine System State Tracking and Response Generation}
In contrast to coarse system state tracking, fine system state tracking involves identifying the exact phrase that answers the query from a ranked subset. The extracted phrase is converted into a natural language response that is coherent within the context of the conversation.

\mbox{cTBLS} employs GPT-3.5 \cite{brown2020language} to perform fine system state tracking and response generation jointly. GPT-3.5 is prompted to generate a natural language response to a follow-up query conditioned on cells of the table ranked by their relevance to the query as obtained from the coarse state tracker. The prompt includes the dialogue history, ranked knowledge sources, and the query to be answered. The bottom-right section of Figure~\ref{fig:system_overview} outlines this process. 

\section{Experiments}
\label{sec:Experiments}

\subsection{\textsc{HybriDialogue}}
The \textsc{HybriDialogue} dataset \cite{nakamura-etal-2022-hybridialogue} comprises 4800 natural language conversations grounded in text and tabular information from Wikipedia. Crowdsourced workers break down multi-hop questions from the OTT-QA dataset \cite{chen2020open} into natural questions and conversational responses related to tabular data. On average, dialogues in the dataset consist of 4-5 conversation turns, with a total of 21,070 turns available in the dataset. Examples of conversations can be found in Figures \ref{fig:dial_1_example} and \ref{fig:dial_2_example}.

\subsection{Table Retrieval}

The first conversation turn of \textsc{HybriDialogue} requires selecting the correct table based on the input query for which we use the Dense Table Retriever outlined in Section~\ref{subsec:tab_ret}.  The Dense Table Retriever is fine-tuned for 20 epochs using Adam \cite{kingma2014adam} with a learning rate of 1e-6 and a linear learning schedule with five warmup steps. The loss function is a modification of the contrastive loss implementation from ConVIRT \cite{zhang2022contrastive}, with image embeddings replaced by table embeddings. The table retriever used in the \textsc{HybriDialogue} paper \cite{nakamura-etal-2022-hybridialogue} was the BM25Okapi Retriever \cite{trotman2014improvements} from \href{https://github.com/dorianbrown/rank_bm25}{rank-bm25}. According to the results presented in Table \ref{tab:retrieval}, \mbox{cTBLS}-DTR outperforms BM25 in terms of Mean Reciprocal Rank (MRR), Top-1 Accuracy, and Top-3 Accuracy on \textsc{HybriDialogue}.
\begin{table}[t]
    \centering
    \begin{tabularx}{\columnwidth} { 
   >{\centering\arraybackslash}X 
  | >{\centering\arraybackslash}X 
  | >{\centering\arraybackslash}X 
  | >{\centering\arraybackslash}X  }
    \toprule 
          &  MRR $@$10 & Top 1 Acc & Top 3 Acc  \\
          \hline 
         {\small BM25} & 0.491 & 0.345 &  0.460 \\
         {\small cTBLS-}{\tiny DTR} & {\textbf{0.846}} & {\textbf{0.777}} & {\textbf{0.901}}\\
        \bottomrule 
    \end{tabularx}
    \caption{BM25 vs \mbox{cTBLS}-DTR for retrieval on first turn of conversation, results on \textsc{HybriDialogue} testing dataset. \mbox{cTBLS}-DTR obtains up to 125\% relative improvement over sparse table retrieval}
    \label{tab:retrieval}
\end{table}

\subsection{Coarse State Tracking}
Coarse state tracking ranks cells from a table based on their relevance to a query. As before, the dual-encoder coarse state tracker of \mbox{cTBLS} consists of RoBERTa-base fine-tuned using Adam with a learning rate of 1e-6 and a linear learning schedule with five warmup steps. In contrast to table retrieval, the state tracker uses triplet margin loss with a margin of 1.0~(Equation \ref{eq:triplet_loss}) instead of contrastive loss~(Equation \ref{eq:table_ret}). The results, as demonstrated in Table~\ref{tab:state_tracking}, show that fine-tuning RoBERTa-base solely on \textsc{HybriDialogue} surpasses the performance of SentenceBERT \cite{reimers-gurevych-2019-sentence}. Furthermore, it nearly attains the same MRR~$@10$ as TaPas \cite{herzig-etal-2020-tapas}, even without additional table pre-training on the SQA dataset \cite{iyyer-etal-2017-search}.

\begin{table}[t]
    \centering
    \begin{tabular}{c|c}
    \toprule  
          & MRR$@$10 \\
    \hline 
         SentenceBERT \tiny{\cite{reimers-gurevych-2019-sentence}} & 0.603  \\
         TaPas \tiny{\cite{herzig-etal-2020-tapas}}  & \textbf{0.689} \\
         \mbox{cTBLS} - RoBERTa-base & 0.683 \\
    \bottomrule 
    \end{tabular}
    \caption{System state tracking results on \textsc{HybriDialogue}. \mbox{cTBLS} achieves nearly the same Mean Reciprocal Rank (MRR) $@$ 10 as TaPaS, without additional table pre-training on SQA \cite{iyyer-etal-2017-search}}
    \label{tab:state_tracking}
\end{table}

\begin{table}[b]
    \centering
    \begin{tabular}{c|c|c|c}
    \toprule 
             & Top-1 & Top-3 & Top-10 \\
             \hline 
         \small{\mbox{cTBLS} - RoBERTa-base}  & 0.559 & 0.778 & 0.925 \\
    \bottomrule  
    \end{tabular}
    \caption{Top-k accuracy for \mbox{cTBLS} on coarse system state tracking. \mbox{cTBLS} ranks the correct cell as the top reference in 56\% of follow-up queries on \textsc{HybriDialogue}. The correct cell is ranked in the Top-3 and Top-10 retrievals in approximately 78\% and 93\% of conversations, respectively.}
    \label{tab:dst_topk}
\end{table}

\subsection{Fine State Tracking and Response Generation}
\mbox{cTBLS} uses GPT-3.5 (text-davinci-003) with the existing dialogue context, the current query, and the retrieved references from coarse state tracking to obtain a natural language response. Since fine-tuning the best available version of the model is cost prohibitive, we opt to prompt GPT-3.5 to generate responses instead. 

\begin{table*}[t]
    \centering 
    \begin{tabular}{c|c|c|c|c|c|c}
    \toprule 
         Model & TR & KR & RG &  ROUGE-1 & ROUGE-2 & ROUGE-L \\
         \hline 
         - & BM25 & Top-1 & DialoGPT & 0.207 & 0.042 & 0.181 \\
         - & BM25 & Top-3 & DialoGPT & 0.212 & 0.045 & 0.186 \\
         - & BM25 & Top-1 & GPT3.5 & 0.428 & 0.207 & 0.369 \\ 
         - & BM25 & Top-3 & GPT3.5 & 0.475 & 0.242 & 0.413 \\ 
         \midrule 
         - & DTR & Top-1 & DialoGPT & 0.222 & 0.051 & 0.195 \\
         - & DTR & Top-3 & DialoGPT & 0.226 & 0.059 & 0.199 \\
         - & DTR & Top-1 & GPT3.5 &  0.494 & 0.255 & 0.424 \\ 
         - & DTR & Top-3 & GPT3.5
         &  0.560 & 0.295 & 0.479 \\ 
         \midrule 
         \textsc{HybriDialogue} & Gold & Top-1 & DialoGPT & 0.438 & 0.212 & 0.375 \\
         \mbox{cTBLS} NoK & Gold & - & GPT3.5 & 0.487 & 0.229 & 0.422 \\ 
         \mbox{cTBLS} Top-1 & Gold & Top-1 & GPT3.5 & 0.603 & 0.304 & 0.517 \\ 
         \mbox{cTBLS} Top-3 & Gold & Top-3 & GPT3.5 & \textbf{0.642} & \textbf{0.322} & \textbf{0.548} \\ 
    \bottomrule
    \end{tabular}
        \caption{Ablation study on automatic evaluation metrics ROUGE-1, ROUGE-2, and ROUGE-L Precision.    Using Dense Table Retrieval (DTR) improves results over BM25 across Top-1 and Top-3 knowledge for DialoGPT and GPT3.5. 
        Furthermore, using Top-3 knowledge sources results in better results than using only Top-1 knowledge sources for DialoGPT and GPT3.5 using both table retrieval methods.  
        \mbox{cTBLS} No Knowledge (NoK), Top-1 Knowledge, Top-3 Knowledge, and \textsc{HybriDialogue} use ground truth table retrieval. \mbox{cTBLS} exhibits a 2x relative improvement in ROUGE Precision over \textsc{HybriDialogue}. 
        \mbox{TR: Table Retrieval}, \mbox{KR: Knowledge Retrieval}, \mbox{RG: Response Generation}}
    \label{tab:Automatic eval}
\end{table*}

The results presented in Table \ref{tab:dst_topk} demonstrate that the coarse state tracker successfully retrieves the correct cell in approximately 56\% of conversations during inference. Furthermore, it achieves Top-3 and Top-10 retrievals in approximately 78\% and 93\% of conversations, respectively.
Motivated by these results, the fine state tracker of \mbox{cTBLS} is evaluated in two different configurations by prompting GPT-3.5 augmented with the \mbox{Top-1} and Top-3 knowledge references (\mbox{cTBLS} Top-1 and \mbox{cTBLS Top-3}). Due to limits on token length associated with the OpenAI API, we remove stopwords from the knowledge provided in the prompt and do not experiment with Top-10 knowledge augmentation. 

Since LLMs store factual information in their weights \cite{roberts-etal-2020-much, heinzerling-inui-2021-language}, we compare to few-shot prompting (using two examples) with no knowledge sources (\mbox{cTBLS}-NoK).  Furthermore, to enable a meaningful comparison with existing research \cite{nakamura-etal-2022-hybridialogue}, we measure \mbox{cTBLS} against the system proposed by \textsc{HybriDialogue} that utilizes a fine-tuned DialoGPT-medium \cite{zhang2019dialogpt} model augmented with Top-1 knowledge.

Table \ref{tab:Automatic eval} presents ROUGE-1, ROUGE-2, and ROUGE-L precision \cite{lin-2004-rouge} for all models assessed. The results demonstrate that superior downstream performance can be achieved through improvements in table retrieval. Specifically, when keeping the number of knowledge sources constant, we observe an improvement in ROUGE precision scores when transitioning from BM25 to DTR, and from DTR to gold table retrieval. The inclusion of additional knowledge sources leads to an improved n-gram overlap with the ground truth reference, as evidenced by the Top-3 knowledge augmented models outperforming their Top-1 counterparts utilizing the same table retriever, and \mbox{cTBLS Top-1} outperforming the baseline model \mbox{cTBLS} NoK. Moreover, \mbox{cTBLS} Top-3 achieves the best performance across all automatic metrics, suggesting the benefits of splitting knowledge retrieval into coarse and fine state tracking, and utilizing additional knowledge sources. Finally, all three configurations of \mbox{cTBLS} demonstrate superior performance to \textsc{HybriDialogue}.

\subsection{Human Evaluation}
To gain a deeper understanding of \mbox{cTBLS}, we conducted human evaluation using the metrics outlined by \citet{nakamura-etal-2022-hybridialogue}, namely Coherence, Fluency, and Informativeness. For the evaluation of these metrics, we enlisted crowd workers from Amazon Mechanical Turk (AMT) to assess 50\% of the test data. The evaluation process involved a comparison between the responses generated by \textsc{HybriDialogue} and \mbox{cTBLS} Top-3.

In accordance with the methodology delineated in \citet{nakamura-etal-2022-hybridialogue}, Coherence was defined as the degree to which a response continued the conversation in a logically coherent manner based on prior context. Fluency, conversely, was determined by evaluating absence of grammatical and spelling errors, and appropriate use of parts of speech.

To ensure the quality of the evaluated responses, we engaged crowd workers possessing a Masters qualification on AMT and originating from English-speaking countries (USA, Canada, Australia, New Zealand, or Great Britain). Each task required approximately 30 seconds to complete, and workers were remunerated at a rate of \$0.05 per task. Moreover, to minimize bias and guarantee the dependability of the evaluations, we assigned two crowd workers to assess each response, with a response deemed more coherent or fluent only if both evaluations concurred.

The results presented in Table \ref{tab:coherence_fluency} reveal that the responses generated by \mbox{cTBLS} Top-3 were more coherent than those produced by \textsc{HybriDialogue} in 84.2\% of cases and exhibited greater fluency 82.7\% of the time, suggesting that improvements in table retrieval, knowledge retrieval, and response generation lead to better downstream performance.   

\begin{table}[t]
    \centering
    \begin{tabular}{c|c}
     \toprule
     & \mbox{cTBLS} Top-3 vs \textsc{HybriDialogue} \\
     \hline
     Coherence & 0.842 \\ 
     Fluency & 0.827 \\
     \bottomrule
    \end{tabular}
    \caption{Coherence and Fluency - \mbox{cTBLS} Top-3 is more conversationally coherent than the best performing \textsc{HybriDialogue} system 84.2\% of the time and is more fluent 82.7\% of the time.}
    \label{tab:coherence_fluency}
\end{table}

Informativeness represents the accuracy of machine-generated responses when compared to the ground-truth \cite{nakamura-etal-2022-hybridialogue} and serves as a measure of hallucination in LLMs. Hallucinated responses tend to be less informative, deviating significantly from the ground-truth. 

To evaluate informativeness, crowd workers determined whether generated responses were semantically equivalent to the ground truth response. Each response was assessed by two Turkers, and a response was deemed more informative only if there was inter-annotator agreement. The absence of illustrative examples in the prompting process resulted in responses generated by \mbox{cTBLS} Top-1 and \mbox{cTBLS Top-3} being longer than the ground truth response. Consequently, the knowledge-augmented \mbox{cTBLS} responses were considered informative if all the information provided in the ground truth was encapsulated in the model response, even if \mbox{cTBLS} included supplementary information.

The data in Table \ref{tab:GT_same} indicate that \mbox{cTBLS} Top-3 encompasses the same information as the ground truth response 50\% of the time, a higher rate than \mbox{cTBLS} Top-1 at 45.6\%, exemplifying the benefits of partitioning retrieval into coarse and fine state tracking and augmenting with additional knowledge. Based on these findings, we hypothesize that the attention mechanism in decoder models facilitates additional knowledge retrieval. \mbox{cTBLS NoK} generates the correct response 30.6\% of the time, suggesting that \textsc{HybriDialogue} comprises questions and answers predicated on general world knowledge embedded in the weights of LLMs. Responses produced by \textsc{HybriDialogue} are informative in merely 12.4\% of instances.

\begin{table}[t]
    \centering
    \begin{tabular}{c|c}
    \toprule 
          & Informativeness \\
          \hline 
         \textsc{HybriDialogue} & 0.124 \\
         \mbox{cTBLS} - NoK& 0.306 \\
         \mbox{cTBLS} Top-1 &  0.456 \\
         \textbf{\mbox{cTBLS} Top-3} & \textbf{0.500} \\
    \bottomrule
    \end{tabular}
    \caption{Human Evaluation Metrics - Fraction of cases where model response is semantically equivalent to ground truth response. Using more knowledge sources results in responses that are more informative, helping reduce hallucination.}
    \label{tab:GT_same}
\end{table}

\begin{figure*}[t]
    \centering
    \includegraphics[width=0.9\textwidth]{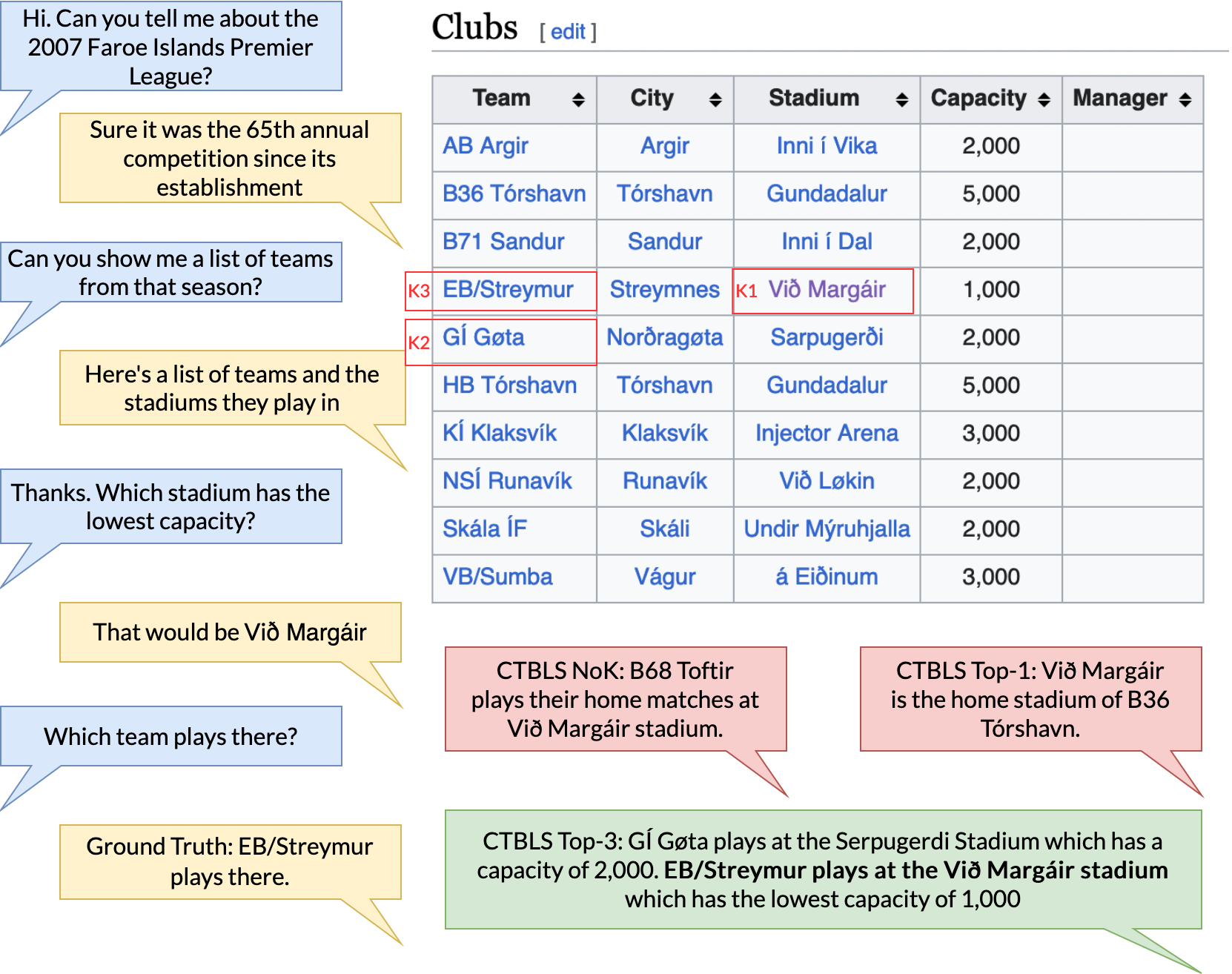}
    \caption{Generated responses vs Ground Truth on \textsc{HybriDialogue} test set. Questions are in \textcolor{MidnightBlue}{blue} and responses in \textcolor{Dandelion}{yellow}. K1, K2, and K3 represent the Top 3 knowledge sources ranked by relevance to the query "Which team plays there?". \mbox{cTBLS} Top-3 is able to leverage K3 to generate the correct response while \mbox{cTBLS} NoK hallucinates a response and \mbox{cTBLS} Top-1 generates an incorrect response based on K1. Table obtained from Wikipedia \href{https://en.wikipedia.org/wiki/2007_Faroe_Islands_Premier_League}{available here}}
    \label{fig:dial_1_example}
\end{figure*}

Figure \ref{fig:dial_1_example} presents a comparison of responses generated by various configurations of \mbox{cTBLS} on the \textsc{HybriDialogue} dataset. The entire dialogue history constitutes the context and is depicted as an exchange between the user (in blue) and the system (in yellow). The final question box represents the follow-up query to be addressed, while the last answer chat box indicates the ground truth response. Knowledge K1, K2, and K3 correspond to cells of the table retrieved during state tracking, based on which responses are produced. \mbox{cTBLS} NoK generates a response solely relying on the context, \mbox{cTBLS} Top-1 formulates a response conditioned on K1, and \mbox{cTBLS} Top-3 devises a response based on K1, K2, and K3.

\begin{figure*}[h!]
    \centering
    \includegraphics[width=\textwidth]{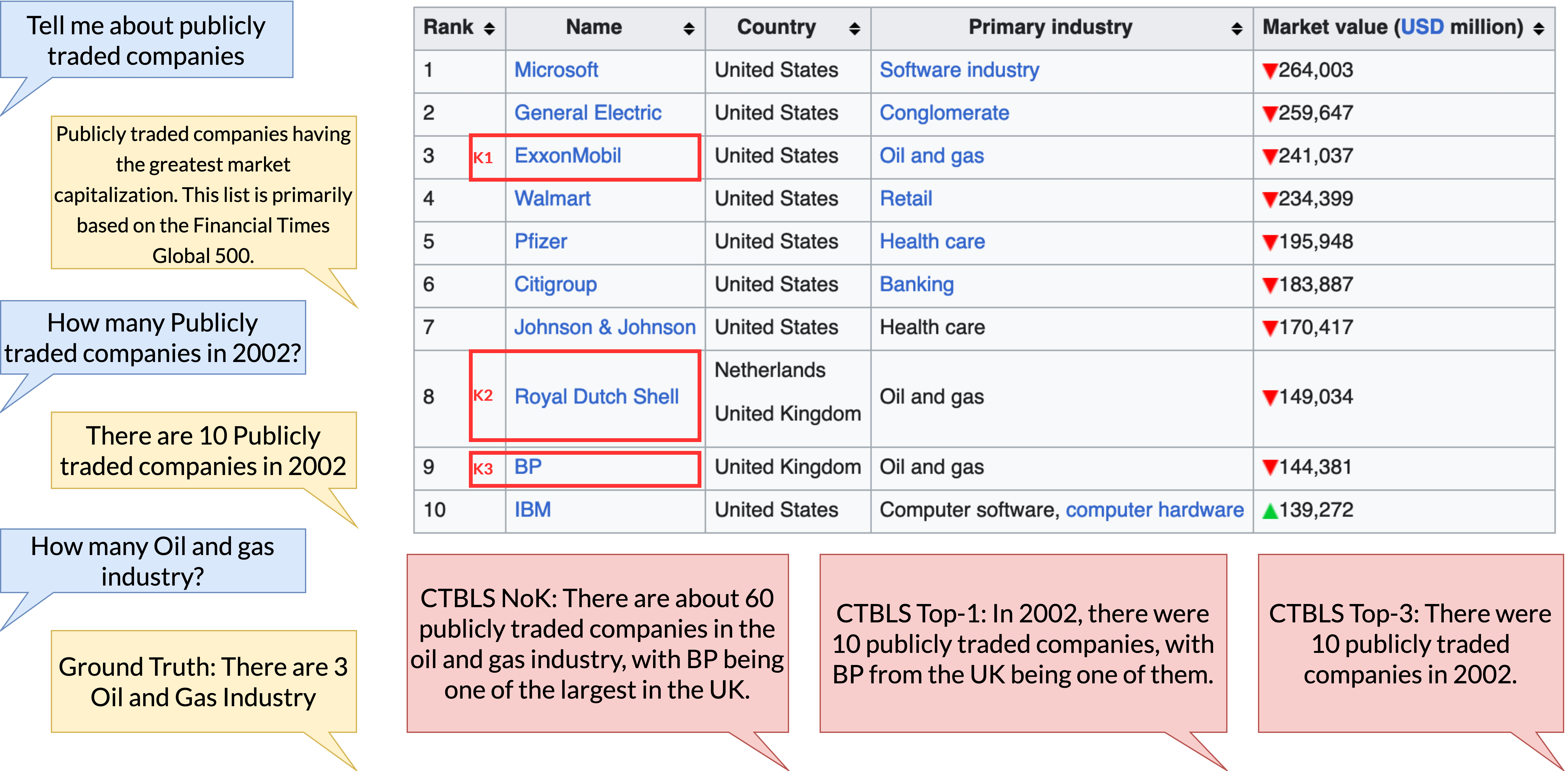}
    \caption{Generated responses vs Ground Truth on \textsc{HybriDialogue} test set. Despite selecting the rows of the table corresponding to Oil and gas industries, \mbox{cTBLS} NoK, Top-1, and Top-3 struggle with counting and hallucinate a response. Table obtained from Wikipedia \href{https://en.wikipedia.org/wiki/List_of_public_corporations_by_market_capitalization}{available here}}
    \label{fig:dial_2_example}
\end{figure*}

\mbox{cTBLS} NoK creates a hallucinated response, answering with the random Faroese club B68 Toftir. Similarly, \mbox{cTBLS} Top-1 hallucinates a response, opting for B36 Tórshavn, as K1 refers to the stadium Viò Margáir rather than the correct club's name. In contrast, \mbox{cTBLS} Top-3 produces the accurate response, EB/Streymur, since K3 contains the necessary information. This example demonstrates the benefits of augmenting response generation with additional pertinent knowledge, which aids in mitigating the hallucination problem \cite{maynez-etal-2020-faithfulness}.

\section{Conclusion}
\label{sec:Conclusion}
In this paper, we introduce Conversational Tables (\mbox{cTBLS}), a system designed to address multi-turn dialogues that are grounded in tabular data. \mbox{cTBLS} separates tabular dialogue into three distinct tasks, specifically table retrieval, system state tracking, and response generation. The dense table retrieval system of \mbox{cTBLS} yields an enhancement of up to 125\% relative to keyword-matching based techniques on the \textsc{HybriDialogue} dataset, with regard to Top-1 Accuracy and Mean Reciprocal Rank~$@$ 10. Furthermore, \mbox{cTBLS} conducts system state tracking utilizing a two-step process shared between encoder and decoder models. This methodology results in natural language responses exhibiting a 2x relative improvement in ROUGE scores. Human evaluators favor \mbox{cTBLS}  +80\% of the time (coherency and fluency) and judge informativeness to be 4x better than the previous state-of-the-art.

\section{Limitations}
Although \mbox{cTBLS} enhances LLMs with tabular knowledge to generate grounded responses, certain limitations remain to be addressed.

Firstly, the efficacy of \mbox{cTBLS} is constrained by the total number of knowledge sources employed during the augmentation process. Token length restrictions in the OpenAI API limit the knowledge augmentation to the top three cells of the table. Another limitation is the incapacity of \mbox{cTBLS} to handle queries pertaining to the entire table. Figure~\ref{fig:dial_2_example} demonstrates one such instance in which the state tracker module accurately retrieves three rows of the table corresponding to oil and gas industries, yet the response generation module fails to utilize this information when transforming the retrieved state into a response. Generally, \mbox{cTBLS} encounters difficulties with counting, comparing the values of cells, and other mathematical operations, an issue we aim to address in future research.

\section{Acknowledgements}
We would like to thank Christopher Richardson, Benjamin Z Reichman, Atishay Jain, and Srikar Bhumireddy for their contributions. We would also like to thank the review committee for their feedback. 
This work was supported by NSF IIS-2112633 and by CoCoSys, one of seven centers in JUMP 2.0, a Semiconductor Research Corporation (SRC) program sponsored by DARPA.


 \bibliography{anthology,custom}

\begin{thebibliography}{55}
\expandafter\ifx\csname natexlab\endcsname\relax\def\natexlab#1{#1}\fi

\bibitem[{Ainslie et~al.(2020)Ainslie, Ontanon, Alberti, Cvicek, Fisher, Pham,
  Ravula, Sanghai, Wang, and Yang}]{ainslie-etal-2020-etc}
Joshua Ainslie, Santiago Ontanon, Chris Alberti, Vaclav Cvicek, Zachary Fisher,
  Philip Pham, Anirudh Ravula, Sumit Sanghai, Qifan Wang, and Li~Yang. 2020.
\newblock \href {https://doi.org/10.18653/v1/2020.emnlp-main.19} {{ETC}:
  Encoding long and structured inputs in transformers}.
\newblock In \emph{Proceedings of the 2020 Conference on Empirical Methods in
  Natural Language Processing (EMNLP)}, pages 268--284, Online. Association for
  Computational Linguistics.

\bibitem[{Bonifacio et~al.(2022)Bonifacio, Abonizio, Fadaee, and
  Nogueira}]{bonifacio2022inpars}
Luiz Bonifacio, Hugo Abonizio, Marzieh Fadaee, and Rodrigo Nogueira. 2022.
\newblock Inpars: Data augmentation for information retrieval using large
  language models.
\newblock \emph{arXiv preprint arXiv:2202.05144}.

\bibitem[{Brown et~al.(2020)Brown, Mann, Ryder, Subbiah, Kaplan, Dhariwal,
  Neelakantan, Shyam, Sastry, Askell et~al.}]{brown2020language}
Tom Brown, Benjamin Mann, Nick Ryder, Melanie Subbiah, Jared~D Kaplan, Prafulla
  Dhariwal, Arvind Neelakantan, Pranav Shyam, Girish Sastry, Amanda Askell,
  et~al. 2020.
\newblock Language models are few-shot learners.
\newblock \emph{Advances in neural information processing systems},
  33:1877--1901.

\bibitem[{Chen(2022)}]{chen2022large}
Wenhu Chen. 2022.
\newblock Large language models are few (1)-shot table reasoners.
\newblock \emph{arXiv preprint arXiv:2210.06710}.

\bibitem[{Chen et~al.(2020{\natexlab{a}})Chen, Chang, Schlinger, Wang, and
  Cohen}]{chen2020open}
Wenhu Chen, Ming-Wei Chang, Eva Schlinger, William Wang, and William~W Cohen.
  2020{\natexlab{a}}.
\newblock Open question answering over tables and text.
\newblock \emph{arXiv preprint arXiv:2010.10439}.

\bibitem[{Chen et~al.(2020{\natexlab{b}})Chen, Su, Yan, and
  Wang}]{chen2020kgpt}
Wenhu Chen, Yu~Su, Xifeng Yan, and William~Yang Wang. 2020{\natexlab{b}}.
\newblock Kgpt: Knowledge-grounded pre-training for data-to-text generation.
\newblock \emph{arXiv preprint arXiv:2010.02307}.

\bibitem[{Chen et~al.(2020{\natexlab{c}})Chen, Zha, Chen, Xiong, Wang, and
  Wang}]{chen-etal-2020-hybridqa}
Wenhu Chen, Hanwen Zha, Zhiyu Chen, Wenhan Xiong, Hong Wang, and William~Yang
  Wang. 2020{\natexlab{c}}.
\newblock \href {https://doi.org/10.18653/v1/2020.findings-emnlp.91}
  {{H}ybrid{QA}: A dataset of multi-hop question answering over tabular and
  textual data}.
\newblock In \emph{Findings of the Association for Computational Linguistics:
  EMNLP 2020}, pages 1026--1036, Online. Association for Computational
  Linguistics.

\bibitem[{Cho et~al.(2018)Cho, Amplayo, Hwang, and Park}]{cho2018adversarial}
Minseok Cho, Reinald~Kim Amplayo, Seung-won Hwang, and Jonghyuck Park. 2018.
\newblock Adversarial tableqa: Attention supervision for question answering on
  tables.
\newblock In \emph{Asian Conference on Machine Learning}, pages 391--406. PMLR.

\bibitem[{Colon-Hernandez et~al.(2021)Colon-Hernandez, Havasi, Alonso, Huggins,
  and Breazeal}]{colon2021combining}
Pedro Colon-Hernandez, Catherine Havasi, Jason Alonso, Matthew Huggins, and
  Cynthia Breazeal. 2021.
\newblock Combining pre-trained language models and structured knowledge.
\newblock \emph{arXiv preprint arXiv:2101.12294}.

\bibitem[{Devlin et~al.(2019)Devlin, Chang, Lee, and
  Toutanova}]{devlin-etal-2019-bert}
Jacob Devlin, Ming-Wei Chang, Kenton Lee, and Kristina Toutanova. 2019.
\newblock \href {https://doi.org/10.18653/v1/N19-1423} {{BERT}: Pre-training of
  deep bidirectional transformers for language understanding}.
\newblock In \emph{Proceedings of the 2019 Conference of the North {A}merican
  Chapter of the Association for Computational Linguistics: Human Language
  Technologies, Volume 1 (Long and Short Papers)}, pages 4171--4186,
  Minneapolis, Minnesota. Association for Computational Linguistics.

\bibitem[{Dinan et~al.(2018)Dinan, Roller, Shuster, Fan, Auli, and
  Weston}]{dinan2018wizard}
Emily Dinan, Stephen Roller, Kurt Shuster, Angela Fan, Michael Auli, and Jason
  Weston. 2018.
\newblock Wizard of wikipedia: Knowledge-powered conversational agents.
\newblock \emph{arXiv preprint arXiv:1811.01241}.

\bibitem[{Eisenschlos et~al.(2021)Eisenschlos, Gor, M{\"u}ller, and
  Cohen}]{eisenschlos2021mate}
Julian~Martin Eisenschlos, Maharshi Gor, Thomas M{\"u}ller, and William~W
  Cohen. 2021.
\newblock Mate: multi-view attention for table transformer efficiency.
\newblock \emph{arXiv preprint arXiv:2109.04312}.

\bibitem[{Hakkani-T\"{u}r et~al.(2014)Hakkani-T\"{u}r, Slaney, Celikyilmaz, and
  Heck}]{Tur2014eyegaze}
Dilek Hakkani-T\"{u}r, Malcolm Slaney, Asli Celikyilmaz, and Larry Heck. 2014.
\newblock \href {https://doi.org/10.1145/2663204.2663277} {Eye gaze for spoken
  language understanding in multi-modal conversational interactions}.
\newblock In \emph{Proceedings of the 16th International Conference on
  Multimodal Interaction}, ICMI '14, page 263–266, New York, NY, USA.
  Association for Computing Machinery.

\bibitem[{Hakkani-Tür et~al.(2014)Hakkani-Tür, Celikyilmaz, Heck, Tur, and
  Zweig}]{hakkani-tr2014probabilistic}
Dilek Hakkani-Tür, Asli Celikyilmaz, Larry Heck, Gokhan Tur, and Geoff Zweig.
  2014.
\newblock \href
  {https://www.microsoft.com/en-us/research/publication/probabilistic-enrichment-of-knowledge-graph-entities-for-relation-detection-in-conversational-understanding/}
  {Probabilistic enrichment of knowledge graph entities for relation detection
  in conversational understanding}.
\newblock In \emph{Proceedings of Interspeech}.

\bibitem[{Hannan et~al.(2020)Hannan, Jain, and Bansal}]{hannan2020manymodalqa}
Darryl Hannan, Akshay Jain, and Mohit Bansal. 2020.
\newblock Manymodalqa: Modality disambiguation and qa over diverse inputs.
\newblock In \emph{Proceedings of the AAAI Conference on Artificial
  Intelligence}, volume~34, pages 7879--7886.

\bibitem[{Heck et~al.(2013)Heck, Hakkani-T{\"u}r, Chinthakunta, Tur, Iyer,
  Parthasacarthy, Stifelman, Shriberg, and Fidler}]{heck2013multimodal}
Larry Heck, Dilek Hakkani-T{\"u}r, Madhu Chinthakunta, Gokhan Tur, Rukmini
  Iyer, Partha Parthasacarthy, Lisa Stifelman, Elizabeth Shriberg, and Ashley
  Fidler. 2013.
\newblock Multimodal conversational search and browse.
\newblock \emph{First Workshop on Speech, Language and Audio in Multimedia
  Marseille, France}.

\bibitem[{Heck and Heck(2020)}]{heck2020zero}
Larry Heck and Simon Heck. 2020.
\newblock Zero-shot visual slot filling as question answering.
\newblock \emph{arXiv preprint arXiv:2011.12340}.

\bibitem[{Heinzerling and Inui(2021)}]{heinzerling-inui-2021-language}
Benjamin Heinzerling and Kentaro Inui. 2021.
\newblock \href {https://doi.org/10.18653/v1/2021.eacl-main.153} {Language
  models as knowledge bases: On entity representations, storage capacity, and
  paraphrased queries}.
\newblock In \emph{Proceedings of the 16th Conference of the European Chapter
  of the Association for Computational Linguistics: Main Volume}, pages
  1772--1791, Online. Association for Computational Linguistics.

\bibitem[{Herzig et~al.(2021)Herzig, M{\"u}ller, Krichene, and
  Eisenschlos}]{herzig-etal-2021-open}
Jonathan Herzig, Thomas M{\"u}ller, Syrine Krichene, and Julian Eisenschlos.
  2021.
\newblock \href {https://doi.org/10.18653/v1/2021.naacl-main.43} {Open domain
  question answering over tables via dense retrieval}.
\newblock In \emph{Proceedings of the 2021 Conference of the North American
  Chapter of the Association for Computational Linguistics: Human Language
  Technologies}, pages 512--519, Online. Association for Computational
  Linguistics.

\bibitem[{Herzig et~al.(2020)Herzig, Nowak, M{\"u}ller, Piccinno, and
  Eisenschlos}]{herzig-etal-2020-tapas}
Jonathan Herzig, Pawel~Krzysztof Nowak, Thomas M{\"u}ller, Francesco Piccinno,
  and Julian Eisenschlos. 2020.
\newblock \href {https://doi.org/10.18653/v1/2020.acl-main.398} {{T}a{P}as:
  Weakly supervised table parsing via pre-training}.
\newblock In \emph{Proceedings of the 58th Annual Meeting of the Association
  for Computational Linguistics}, pages 4320--4333, Online. Association for
  Computational Linguistics.

\bibitem[{Huang et~al.(2015)Huang, Heck, and Ji}]{huang2015leveraging}
Hongzhao Huang, Larry Heck, and Heng Ji. 2015.
\newblock Leveraging deep neural networks and knowledge graphs for entity
  disambiguation.
\newblock \emph{arXiv preprint arXiv:1504.07678}.

\bibitem[{Huang et~al.(2022)Huang, Zhong, Liu, Gong, Jiang, and
  Duan}]{huang2022mixed}
Junjie Huang, Wanjun Zhong, Qian Liu, Ming Gong, Daxin Jiang, and Nan Duan.
  2022.
\newblock Mixed-modality representation learning and pre-training for joint
  table-and-text retrieval in openqa.
\newblock \emph{arXiv preprint arXiv:2210.05197}.

\bibitem[{Huang et~al.(2013)Huang, He, Gao, Deng, Acero, and
  Heck}]{huang2013learning}
Po-Sen Huang, Xiaodong He, Jianfeng Gao, Li~Deng, Alex Acero, and Larry Heck.
  2013.
\newblock Learning deep structured semantic models for web search using
  clickthrough data.
\newblock In \emph{Proceedings of the 22nd ACM international conference on
  Information \& Knowledge Management}, pages 2333--2338. ACM.

\bibitem[{Iyyer et~al.(2017)Iyyer, Yih, and Chang}]{iyyer-etal-2017-search}
Mohit Iyyer, Wen-tau Yih, and Ming-Wei Chang. 2017.
\newblock \href {https://doi.org/10.18653/v1/P17-1167} {Search-based neural
  structured learning for sequential question answering}.
\newblock In \emph{Proceedings of the 55th Annual Meeting of the Association
  for Computational Linguistics (Volume 1: Long Papers)}, pages 1821--1831,
  Vancouver, Canada. Association for Computational Linguistics.

\bibitem[{Izacard and Grave(2021)}]{izacard-grave-2021-leveraging}
Gautier Izacard and Edouard Grave. 2021.
\newblock \href {https://doi.org/10.18653/v1/2021.eacl-main.74} {Leveraging
  passage retrieval with generative models for open domain question answering}.
\newblock In \emph{Proceedings of the 16th Conference of the European Chapter
  of the Association for Computational Linguistics: Main Volume}, pages
  874--880, Online. Association for Computational Linguistics.

\bibitem[{Jauhar et~al.(2016)Jauhar, Turney, and
  Hovy}]{jauhar-etal-2016-tables}
Sujay~Kumar Jauhar, Peter Turney, and Eduard Hovy. 2016.
\newblock \href {https://doi.org/10.18653/v1/P16-1045} {Tables as
  semi-structured knowledge for question answering}.
\newblock In \emph{Proceedings of the 54th Annual Meeting of the Association
  for Computational Linguistics (Volume 1: Long Papers)}, pages 474--483,
  Berlin, Germany. Association for Computational Linguistics.

\bibitem[{Jeronymo et~al.(2023)Jeronymo, Bonifacio, Abonizio, Fadaee, Lotufo,
  Zavrel, and Nogueira}]{jeronymo2023inpars}
Vitor Jeronymo, Luiz Bonifacio, Hugo Abonizio, Marzieh Fadaee, Roberto Lotufo,
  Jakub Zavrel, and Rodrigo Nogueira. 2023.
\newblock Inpars-v2: Large language models as efficient dataset generators for
  information retrieval.
\newblock \emph{arXiv preprint arXiv:2301.01820}.

\bibitem[{Jia et~al.(2017)Jia, Heck, Hakkani-Tür, and
  Nikolov}]{jia2017learning}
Robin Jia, Larry Heck, Dilek Hakkani-Tür, and Georgi Nikolov. 2017.
\newblock \href {https://doi.org/10.1109/ICASSP.2017.7953253} {Learning
  concepts through conversations in spoken dialogue systems}.
\newblock In \emph{2017 IEEE International Conference on Acoustics, Speech and
  Signal Processing (ICASSP)}, pages 5725--5729.

\bibitem[{Jin et~al.(2022)Jin, Siebert, Li, and Chen}]{jin2022surveytable}
Nengzheng Jin, Joanna Siebert, Dongfang Li, and Qingcai Chen. 2022.
\newblock A survey on table question answering: Recent advances.
\newblock In \emph{Knowledge Graph and Semantic Computing: Knowledge Graph
  Empowers the Digital Economy}, pages 174--186, Singapore. Springer Nature
  Singapore.

\bibitem[{Karpukhin et~al.(2020)Karpukhin, Oguz, Min, Lewis, Wu, Edunov, Chen,
  and Yih}]{karpukhin-etal-2020-dense}
Vladimir Karpukhin, Barlas Oguz, Sewon Min, Patrick Lewis, Ledell Wu, Sergey
  Edunov, Danqi Chen, and Wen-tau Yih. 2020.
\newblock \href {https://doi.org/10.18653/v1/2020.emnlp-main.550} {Dense
  passage retrieval for open-domain question answering}.
\newblock In \emph{Proceedings of the 2020 Conference on Empirical Methods in
  Natural Language Processing (EMNLP)}, pages 6769--6781, Online. Association
  for Computational Linguistics.

\bibitem[{Kingma and Ba(2014)}]{kingma2014adam}
Diederik~P Kingma and Jimmy Ba. 2014.
\newblock Adam: A method for stochastic optimization.
\newblock \emph{arXiv preprint arXiv:1412.6980}.

\bibitem[{Lewis et~al.(2020)Lewis, Perez, Piktus, Petroni, Karpukhin, Goyal,
  K\"{u}ttler, Lewis, Yih, Rockt\"{a}schel, Riedel, and
  Kiela}]{lewis2020retrieval}
Patrick Lewis, Ethan Perez, Aleksandra Piktus, Fabio Petroni, Vladimir
  Karpukhin, Naman Goyal, Heinrich K\"{u}ttler, Mike Lewis, Wen-tau Yih, Tim
  Rockt\"{a}schel, Sebastian Riedel, and Douwe Kiela. 2020.
\newblock \href
  {https://proceedings.neurips.cc/paper/2020/file/6b493230205f780e1bc26945df7481e5-Paper.pdf}
  {Retrieval-augmented generation for knowledge-intensive nlp tasks}.
\newblock In \emph{Advances in Neural Information Processing Systems},
  volume~33, pages 9459--9474. Curran Associates, Inc.

\bibitem[{Li et~al.(2021)Li, Ng, Xu, Zhu, Wang, and Xiang}]{li-etal-2021-dual}
Alexander~Hanbo Li, Patrick Ng, Peng Xu, Henghui Zhu, Zhiguo Wang, and Bing
  Xiang. 2021.
\newblock \href {https://doi.org/10.18653/v1/2021.acl-long.315} {Dual
  reader-parser on hybrid textual and tabular evidence for open domain question
  answering}.
\newblock In \emph{Proceedings of the 59th Annual Meeting of the Association
  for Computational Linguistics and the 11th International Joint Conference on
  Natural Language Processing (Volume 1: Long Papers)}, pages 4078--4088,
  Online. Association for Computational Linguistics.

\bibitem[{Lin(2004)}]{lin-2004-rouge}
Chin-Yew Lin. 2004.
\newblock \href {https://aclanthology.org/W04-1013} {{ROUGE}: A package for
  automatic evaluation of summaries}.
\newblock In \emph{Text Summarization Branches Out}, pages 74--81, Barcelona,
  Spain. Association for Computational Linguistics.

\bibitem[{Liu et~al.(2019)Liu, Ott, Goyal, Du, Joshi, Chen, Levy, Lewis,
  Zettlemoyer, and Stoyanov}]{liu2019roberta}
Yinhan Liu, Myle Ott, Naman Goyal, Jingfei Du, Mandar Joshi, Danqi Chen, Omer
  Levy, Mike Lewis, Luke Zettlemoyer, and Veselin Stoyanov. 2019.
\newblock Roberta: A robustly optimized bert pretraining approach.
\newblock \emph{arXiv preprint arXiv:1907.11692}.

\bibitem[{Maynez et~al.(2020)Maynez, Narayan, Bohnet, and
  McDonald}]{maynez-etal-2020-faithfulness}
Joshua Maynez, Shashi Narayan, Bernd Bohnet, and Ryan McDonald. 2020.
\newblock \href {https://doi.org/10.18653/v1/2020.acl-main.173} {On
  faithfulness and factuality in abstractive summarization}.
\newblock In \emph{Proceedings of the 58th Annual Meeting of the Association
  for Computational Linguistics}, pages 1906--1919, Online. Association for
  Computational Linguistics.

\bibitem[{Moiseev et~al.(2022)Moiseev, Dong, Alfonseca, and
  Jaggi}]{moiseev-etal-2022-skill}
Fedor Moiseev, Zhe Dong, Enrique Alfonseca, and Martin Jaggi. 2022.
\newblock \href {https://doi.org/10.18653/v1/2022.naacl-main.113} {{SKILL}:
  Structured knowledge infusion for large language models}.
\newblock In \emph{Proceedings of the 2022 Conference of the North American
  Chapter of the Association for Computational Linguistics: Human Language
  Technologies}, pages 1581--1588, Seattle, United States. Association for
  Computational Linguistics.

\bibitem[{Nakamura et~al.(2022)Nakamura, Levy, Tuan, Chen, and
  Wang}]{nakamura-etal-2022-hybridialogue}
Kai Nakamura, Sharon Levy, Yi-Lin Tuan, Wenhu Chen, and William~Yang Wang.
  2022.
\newblock \href {https://doi.org/10.18653/v1/2022.findings-acl.41}
  {{H}ybri{D}ialogue: An information-seeking dialogue dataset grounded on
  tabular and textual data}.
\newblock In \emph{Findings of the Association for Computational Linguistics:
  ACL 2022}, pages 481--492, Dublin, Ireland. Association for Computational
  Linguistics.

\bibitem[{Pasupat and Liang(2015)}]{pasupat-liang-2015-compositional}
Panupong Pasupat and Percy Liang. 2015.
\newblock \href {https://doi.org/10.3115/v1/P15-1142} {Compositional semantic
  parsing on semi-structured tables}.
\newblock In \emph{Proceedings of the 53rd Annual Meeting of the Association
  for Computational Linguistics and the 7th International Joint Conference on
  Natural Language Processing (Volume 1: Long Papers)}, pages 1470--1480,
  Beijing, China. Association for Computational Linguistics.

\bibitem[{Peng et~al.(2023)Peng, Galley, He, Cheng, Xie, Hu, Huang, Liden, Yu,
  Chen, and Gao}]{peng2023check}
Baolin Peng, Michel Galley, Pengcheng He, Hao Cheng, Yujia Xie, Yu~Hu, Qiuyuan
  Huang, Lars Liden, Zhou Yu, Weizhu Chen, and Jianfeng Gao. 2023.
\newblock \href {http://arxiv.org/abs/2302.12813} {Check your facts and try
  again: Improving large language models with external knowledge and automated
  feedback}.

\bibitem[{Reimers and Gurevych(2019)}]{reimers-gurevych-2019-sentence}
Nils Reimers and Iryna Gurevych. 2019.
\newblock \href {https://doi.org/10.18653/v1/D19-1410} {Sentence-{BERT}:
  Sentence embeddings using {S}iamese {BERT}-networks}.
\newblock In \emph{Proceedings of the 2019 Conference on Empirical Methods in
  Natural Language Processing and the 9th International Joint Conference on
  Natural Language Processing (EMNLP-IJCNLP)}, pages 3982--3992, Hong Kong,
  China. Association for Computational Linguistics.

\bibitem[{Richardson and Heck(2023)}]{richardson2023commonsense}
Christopher Richardson and Larry Heck. 2023.
\newblock Commonsense reasoning for conversational ai: A survey of the state of
  the art.
\newblock \emph{Workshop on Knowledge Augmented Methods for NLP,
  (KnowledgeNLP-AAAI’23)}.

\bibitem[{Roberts et~al.(2020)Roberts, Raffel, and
  Shazeer}]{roberts-etal-2020-much}
Adam Roberts, Colin Raffel, and Noam Shazeer. 2020.
\newblock \href {https://doi.org/10.18653/v1/2020.emnlp-main.437} {How much
  knowledge can you pack into the parameters of a language model?}
\newblock In \emph{Proceedings of the 2020 Conference on Empirical Methods in
  Natural Language Processing (EMNLP)}, pages 5418--5426, Online. Association
  for Computational Linguistics.

\bibitem[{Shi et~al.(2023)Shi, Min, Yasunaga, Seo, James, Lewis, Zettlemoyer,
  and Yih}]{shi2023replug}
Weijia Shi, Sewon Min, Michihiro Yasunaga, Minjoon Seo, Rich James, Mike Lewis,
  Luke Zettlemoyer, and Wen-tau Yih. 2023.
\newblock Replug: Retrieval-augmented black-box language models.
\newblock \emph{arXiv preprint arXiv:2301.12652}.

\bibitem[{Shuster et~al.(2021)Shuster, Poff, Chen, Kiela, and
  Weston}]{shuster-etal-2021-retrieval-augmentation}
Kurt Shuster, Spencer Poff, Moya Chen, Douwe Kiela, and Jason Weston. 2021.
\newblock \href {https://doi.org/10.18653/v1/2021.findings-emnlp.320}
  {Retrieval augmentation reduces hallucination in conversation}.
\newblock In \emph{Findings of the Association for Computational Linguistics:
  EMNLP 2021}, pages 3784--3803, Punta Cana, Dominican Republic. Association
  for Computational Linguistics.

\bibitem[{Sundar and Heck(2022)}]{sundar-heck-2022-multimodal}
Anirudh Sundar and Larry Heck. 2022.
\newblock \href {https://doi.org/10.18653/v1/2022.nlp4convai-1.12} {Multimodal
  conversational {AI}: A survey of datasets and approaches}.
\newblock In \emph{Proceedings of the 4th Workshop on NLP for Conversational
  AI}, pages 131--147, Dublin, Ireland. Association for Computational
  Linguistics.

\bibitem[{Trotman et~al.(2014)Trotman, Puurula, and
  Burgess}]{trotman2014improvements}
Andrew Trotman, Antti Puurula, and Blake Burgess. 2014.
\newblock \href {https://doi.org/10.1145/2682862.2682863} {Improvements to bm25
  and language models examined}.
\newblock In \emph{Proceedings of the 2014 Australasian Document Computing
  Symposium}, ADCS '14, page 58–65, New York, NY, USA. Association for
  Computing Machinery.

\bibitem[{Vaswani et~al.(2017)Vaswani, Shazeer, Parmar, Uszkoreit, Jones,
  Gomez, Kaiser, and Polosukhin}]{vaswani2017attention}
Ashish Vaswani, Noam Shazeer, Niki Parmar, Jakob Uszkoreit, Llion Jones,
  Aidan~N Gomez, {\L}ukasz Kaiser, and Illia Polosukhin. 2017.
\newblock Attention is all you need.
\newblock \emph{Advances in neural information processing systems}, 30.

\bibitem[{Yang et~al.(2022)Yang, Gupta, Upadhyay, He, Goel, and
  Paul}]{yang2022tableformer}
Jingfeng Yang, Aditya Gupta, Shyam Upadhyay, Luheng He, Rahul Goel, and Shachi
  Paul. 2022.
\newblock Tableformer: Robust transformer modeling for table-text encoding.
\newblock \emph{arXiv preprint arXiv:2203.00274}.

\bibitem[{Yin et~al.(2020)Yin, Neubig, Yih, and Riedel}]{yin-etal-2020-tabert}
Pengcheng Yin, Graham Neubig, Wen-tau Yih, and Sebastian Riedel. 2020.
\newblock \href {https://doi.org/10.18653/v1/2020.acl-main.745} {{T}a{BERT}:
  Pretraining for joint understanding of textual and tabular data}.
\newblock In \emph{Proceedings of the 58th Annual Meeting of the Association
  for Computational Linguistics}, pages 8413--8426, Online. Association for
  Computational Linguistics.

\bibitem[{Yu et~al.(2022)Yu, Iter, Wang, Xu, Ju, Sanyal, Zhu, Zeng, and
  Jiang}]{yu2022generate}
Wenhao Yu, Dan Iter, Shuohang Wang, Yichong Xu, Mingxuan Ju, Soumya Sanyal,
  Chenguang Zhu, Michael Zeng, and Meng Jiang. 2022.
\newblock Generate rather than retrieve: Large language models are strong
  context generators.
\newblock \emph{arXiv preprint arXiv:2209.10063}.

\bibitem[{Zayats et~al.(2021)Zayats, Toutanova, and
  Ostendorf}]{zayats-etal-2021-representations}
Vicky Zayats, Kristina Toutanova, and Mari Ostendorf. 2021.
\newblock \href {https://doi.org/10.18653/v1/2021.eacl-main.253}
  {Representations for question answering from documents with tables and text}.
\newblock In \emph{Proceedings of the 16th Conference of the European Chapter
  of the Association for Computational Linguistics: Main Volume}, pages
  2895--2906, Online. Association for Computational Linguistics.

\bibitem[{Zhang et~al.(2019)Zhang, Sun, Galley, Chen, Brockett, Gao, Gao, Liu,
  and Dolan}]{zhang2019dialogpt}
Yizhe Zhang, Siqi Sun, Michel Galley, Yen-Chun Chen, Chris Brockett, Xiang Gao,
  Jianfeng Gao, JJ~(Jingjing) Liu, and Bill Dolan. 2019.
\newblock \href
  {https://www.microsoft.com/en-us/research/publication/dialogpt-large-scale-generative-pre-training-for-conversational-response-generation/}
  {Dialogpt: Large-scale generative pre-training for conversational response
  generation}.
\newblock In \emph{arXiv:1911.00536}.

\bibitem[{Zhang et~al.(2022)Zhang, Jiang, Miura, Manning, and
  Langlotz}]{zhang2022contrastive}
Yuhao Zhang, Hang Jiang, Yasuhide Miura, Christopher~D Manning, and Curtis~P
  Langlotz. 2022.
\newblock Contrastive learning of medical visual representations from paired
  images and text.
\newblock In \emph{Machine Learning for Healthcare Conference}, pages 2--25.
  PMLR.

\bibitem[{Zhao et~al.(2022)Zhao, Li, Li, and
  Zhang}]{zhao-etal-2022-multihiertt}
Yilun Zhao, Yunxiang Li, Chenying Li, and Rui Zhang. 2022.
\newblock \href {https://doi.org/10.18653/v1/2022.acl-long.454}
  {{M}ulti{H}iertt: Numerical reasoning over multi hierarchical tabular and
  textual data}.
\newblock In \emph{Proceedings of the 60th Annual Meeting of the Association
  for Computational Linguistics (Volume 1: Long Papers)}, pages 6588--6600,
  Dublin, Ireland. Association for Computational Linguistics.

\end{thebibliography}
\bibliographystyle{acl_natbib}

\end{document}